# Unsupervised detection of diachronic word sense evolution

**Jean-François Delpech**
WebGlyphs, Inc.
1515 N. Colonial Ct
Arlington, VA 22209-1439
United States
jfdelpech.webglyphs@gmail.com

Most words have several senses and connotations which evolve in time due to semantic shift, so that closely related words may gain different or even opposite meanings over the years. This evolution is very relevant to the study of language and of cultural changes, but the tools currently available for diachronic semantic analysis have significant, inherent limitations and are not suitable for real-time analysis. In this article, we demonstrate how the linearity of random vectors techniques enables building time series of congruent word embeddings (or semantic spaces) which can then be compared and combined linearly without loss of precision over any time period to detect diachronic semantic shifts. We show how this approach yields time trajectories of polysemous words such as *amazon* or *apple*, enables following semantic drifts and gender bias across time, reveals the shifting instantiations of stable concepts such as *hurricane* or *president*. This very fast, linear approach can easily be distributed over many processors to follow in real time streams of social media such as Twitter or Facebook; the resulting, time-dependent semantic spaces can then be combined at will by simple additions or subtractions.

## 1. Introduction

Over the last few years, following the work of Mikolov *et al.*[1,2] and other investigators[3,4], successful word representation methods inspired from neural-network language modeling have been demonstrated and explained. These representations are nonlinear[5] and the semantic spaces in which they are embedded are randomly oriented as the cost functions used for training are invariant under rotation[6]. These two properties make it very difficult to combine the semantic spaces obtained from several corpora and thus to compare their semantic neighborhoods: the *word alignment* required for diachronic investigations is thus a difficult problem.

Approximate, time-dependent word embeddings have nevertheless been proposed in a few recent publications, with encouraging results, but they are computationally intensive and have limited time resolutions.

Yao *et al.*[6] show that a low-rank factorization of the pointwise mutual information matrix can be performed by adding to the usual minimization problem a smoothing term to encourage word embeddings to remain aligned between time slices at $t_{i-1}$ and $t_i$; a parameter $\tau$ controls how fast the embedding may change between time slices and thus the time resolution. Obviously, even though alignement may be reasonable between consecutive time slices, it will usually be poor over larger time periods. They use as a dataset articles from the *New York Times* published from January 1990 to July 2016; they have made this corpus public and it was used in the present article[7].

In a recent article, Kulkarni *et al.*[8] discuss three approaches for detecting semantic shifts of words:

- Their first approach involves simply counting the change in the relative number of occurrences (frequency) of a given word between time slices. Despite its simplicity, this method sometimes yields useful results.
- Their second approach involves tracking the functionality of a word; for example, they show that *apple* is almost always a common noun before 1976 but often becomes a proper noun after the creation of the eponymous company.
- Their third approach is more specific, as it actually enables tracking word senses across time slices (or epochs) by first learning embeddings for each epoch and then aligning them. However, their alignment process starts with two assumptions of uncertain validity, that (a) the semantic spaces are equivalent under a linear transformation and that (b) the meaning of most words do not shift over time.

They give a few interesting examples using a 24 month long sample from Twitter from September 2011 to October 2013 (resolution of 1 month), the Amazon Movie Reviews dataset from 2000 to 2012 (resolution of 1 year) and the Google Books Ngram corpus over the last 105 years, with a 5-year resolution.

Hamilton *et al.*[9] have used similar approaches to demonstrate empirically that polysemous words change at a faster rate than non-polysemous words. Also, gender and ethnic stereotyping has been quantified over the past 100 years by Garg *et al.*[10] using word embeddings derived from various sources.

These nonlinear approaches rely on a foundational assumption in Natural Language Processing (NLP), the *distributional hypothesis*[11], according to which linguistic items with similar distributions (i.e. occurring in the same contexts) have similar meanings, where "context" means in practice a moving window of width $w$ or perhaps sometimes a full sentence.

A linear approach is also obviously compatible with this assumption and a very simple, linear embedding transformation [12,13,14] relies on the fact that in $\mathbb{R}^d$ an exponentially large number $N \gg d$ of random vectors quasi-orthogonal to each other[15] can be created. These vectors can then be treated to a good approximation as if they were orthogonal and

  a. to each distinct, significant term $t$ in a large set of documents is associated a normalized random *seed vector* belonging to $\mathbb{R}^d$ which is quasi-orthogonal to any other seed vector, and
  b. to each term $t$ is then attached as *term vector*, i.e. a linear, weighted combination of the seed vectors of the terms co-occurring with $t$ in all windows of fixed size $w$.

With a window size $w = 5$, this means that the sentence fragment centered around term $t_i$

$$\ldots \quad t_{i-3} \quad t_{i-2} \quad t_{i-1} \quad t_i \quad t_{i+1} \quad t_{i+2} \quad t_{i+3} \quad \ldots$$

will increment term vector $|\mathcal{T}_i\rangle$ by the sum of four seed vectors $|\mathcal{T}^s\rangle$:

$$|\mathcal{T}_i\rangle \mathrel{+}= \rho_{i-2}|\mathcal{T}^s_{i-2}\rangle + \rho_{i-1}|\mathcal{T}^s_{i-1}\rangle + \rho_{i+1}|\mathcal{T}^s_{i+1}\rangle + \rho_{i+2}|\mathcal{T}^s_{i+2}\rangle \tag{1}$$

where the $\rho_{i+\delta}$ are multiplicative coefficients, for example weights. If term $t_i$ occurs $N$ times, vector $|\mathcal{T}_i\rangle$ will be the sum of $N$ equations such as Equation 1, with terms which may be different or not, depending on the strength of their association with $t_i$. This is obviously independent of sentence order.

Each term vector $|\mathcal{T}_i\rangle$ as well as any linear combination of term vectors such as documents are themselves embedded in $\mathbb{R}^d$. In this $d$-dimensional Euclidean semantic space noted $\mathcal{S}$, the similarity $\sigma_{ij}$ between terms $t_i$ and $t_j$ is the scalar product of the associated, normalized term vectors:

$$\sigma_{ij} = \langle \mathcal{T}_i | \mathcal{T}_j \rangle$$

(In what follows, inner products will always be assumed to be evaluated between normalized vectors, unless otherwise noted.)

Consider a partition $k$ of corpus $K$; all term vectors $|\mathcal{T}_i^k\rangle$ are embedded in semantic space $\mathcal{S}_k$ and one can define symbolically the semantic space $|\mathcal{S}_K\rangle$ of the whole corpus

$$\mathcal{S}_K = \sum_{k \in K} \mathcal{S}_k \tag{2}$$

since the coordinates of each vector result from a combination of the same linear operations in a different order.

A marked advantage of this linear, transparent process is that it is by nature incremental, which is very important in diachronic studies; a small addition to (or deletion from) the data set involves only a small, finite number of words, at least to a first, very good approximation (over time, word frequencies do vary, and this may be reflected in slowly evolving logarithmic weight factors.) Seed vectors are derived from word strings by a hashing function: they are thus reproducible and semantic spaces from different epochs can be combined at will, as in Equation 2. In fact, despite the use of random vectors, the whole process is totally deterministic in that, even with sentences in random order (provided word order remains fixed within each sentence) the same corpus with the same seed vectors will always generate the same semantic space within rounding errors.

In this work, the embedding space dimensionality was $d = 300$ and the window size was $w = 11$ (5 words before and 5 after the central word.) Yao's *et al.* dataset[7] was used. The 27 time slices contain a total of 92,965 articles and 91,423,989 words; out of a total of 455,008 distinct words, 142,439 distinct words remain after removing the 100 most frequent words (which occur 42,703,951 times, or 47% of the time) and ignoring words occurring less than 5 times. A total semantic space $\mathcal{S}_K$ was also created by combining all 27 partial spaces $\mathcal{S}_k$.

## 2. Linearity

The time evolutions of words *bush* and *enron* are shown in Figure 1. It can be seen that the norm (square of the length of the term vector for each year, in this linear process) and the frequency (number of occurrences) closely track each other, as would be expected. This is quite different from what is shown in Figures 2 and 3 of Yao *et al.*[6]: between 1993 and 1998, they find that the frequencies for *bush* become almost zero but that the nonlinear norms remain substantial, which seems to indicate that the time resolution of their method is of the order of 2 or 3 years.

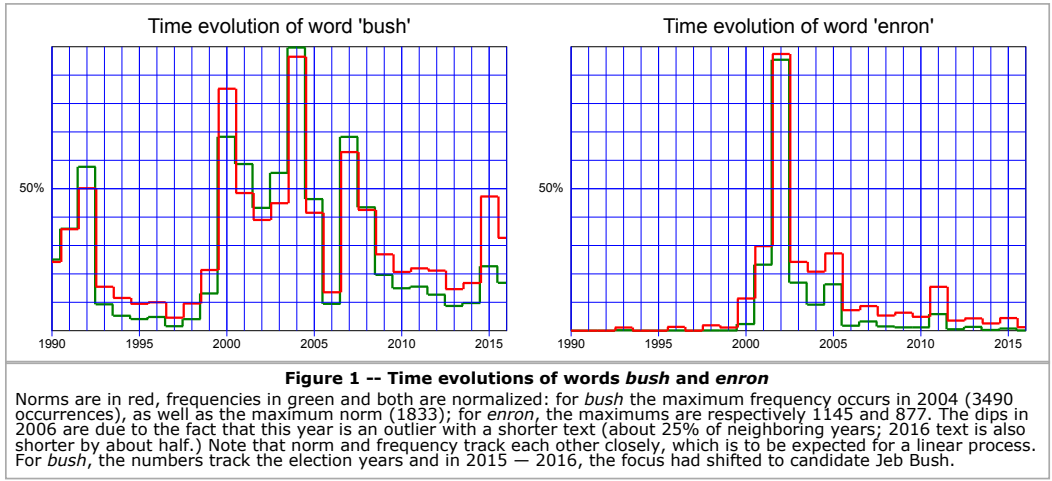

**Figure 1 -- Time evolutions of words *bush* and *enron***
Norms are in red, frequencies in green and both are normalized: for *bush* the maximum frequency occurs in 2004 (3490 occurrences), as well as the maximum norm (1833); for *enron*, the maximums are respectively 1145 and 877. The dips in 2006 are due to the fact that this year is an outlier with a shorter text (about 25% of neighboring years; 2016 text is also shorter by about half.) Note that norm and frequency track each other closely, which is to be expected for a linear process. For *bush*, the numbers track the election years and in 2015 — 2016, the focus had shifted to candidate Jeb Bush.

## 3. Time trajectories

In their figure 1, Yao *et al.*[6] show time trajectories of brand names by plotting for each epoch the 2D t-SNE projections of the brand name and of its closest neighbors; the exact meaning of these plots is not obvious as the contribution of the slice-to-slice alignments and of the projection technique are difficult to unravel.

In a linear, random-vector approach, time evolution can easily be analyzed with the following algorithm:

**Algorithm 1** - Time evolution algorithm for term $t$

1 : Extract and normalize the vector $|\mathcal{T}_t^K\rangle$ of term $t$ from the *total* semantic space $\mathcal{S}_K$
2 : Find a set $R$ of representative neighbors of term $t$ in $\mathcal{S}_K$
3 : **for each epoch** $k \in K$
4 :   Extract and normalize the vector $|\mathcal{T}_r^k\rangle$ of each member $r \in R$ from the *partial* semantic space $\mathcal{S}_k$
5 :   Compute the similarity $\sigma_r^k = \langle \mathcal{T}_r^k | \mathcal{T}_t^K \rangle$ for each member $r$ and keep closest terms
6 : **return** Top neighbors as a function of $k$
7 : Tabulate top neighbors

Note that in this algorithm, a vector extracted from the total semantic space $\mathcal{S}_K$ is compared to vectors extracted from partitions $\mathcal{S}_k$; this does not require any approximation because Equation 2 is strictly valid for a linear process.

All references to *amazon* were close to *rainforest* and to related items before 1998, as shown in Table 1. The company was founded in 1994 by Jeff Bezos; he appears in second place (pink rectangle) in 1997; terms related to *rainforest* being still dominant. After 1997, terms relating to the company and its competitors tend to be more frequent but both senses coexist.

|  | '90 | '91 | '92 | '93 | '94 | '95 | '96 | '97 | '98 | '99 | '00 | '01 | '02 | '03 | '04 | '05 | '06 | '07 | '08 | '09 | '10 | '11 | '12 | '13 | '14 | '15 | '16 |
|---|---|---|---|---|---|---|---|---|---|---|---|---|---|---|---|---|---|---|---|---|---|---|---|---|---|---|---|
| deforestation |  |  | ■ |  |  | ▫ |  |  |  |  |  |  |  |  |  | ▫ |  |  |  |  |  |  |  |  |  |  |  |
| rainforest | ■ |  |  |  |  |  |  | ■ |  |  |  |  |  |  |  |  |  |  |  |  |  | ▫ |  |  |  |  |  |
| iquitos | ▫ |  |  |  |  | ■ |  |  |  |  |  |  |  |  |  |  |  |  |  |  |  |  |  |  |  |  |  |
| manaus |  | ■ |  | ■ |  |  | ■ |  | ▫ | ▫ |  |  |  |  |  |  |  |  |  |  |  |  |  |  |  |  |  |
| bezos |  |  |  |  |  |  |  | ▫ | ■ | ■ |  | ▫ |  |  |  | ▫ |  |  | ▫ | ▫ | ▫ |  |  |  |  |  |  |
| streaming |  |  |  | ▫ |  |  |  |  |  |  |  |  |  |  |  |  |  |  |  |  |  | ▫ |  |  | ▫ | ■ | ▫ |
| online |  |  |  |  |  | ■ | ■ | ▫ |  | ▫ | ■ | ■ | ■ | ■ | ■ | ■ |  | ■ | ■ | ■ | ■ | ■ |  |  |  |  | ■ |
| downloads |  |  |  |  |  |  |  |  | ▫ |  |  |  |  |  |  |  | ■ |  |  |  |  |  |  |  |  |  |  |
| netflix |  |  |  |  |  |  |  |  |  |  |  |  |  |  |  |  |  |  |  |  | ▫ |  | ▫ | ■ | ▫ | ▫ |  |
| Nr. of occurrences | 18 | 36 | 7 | 19 | 5 | 1 | 10 | 13 | 26 | 76 | 154 | 75 | 62 | 54 | 70 | 75 | 5 | 73 | 91 | 76 | 71 | 101 | 196 | 102 | 399 | 131 | 83 |

**Table 1 -- Time trajectory of *amazon***
This table shows the closest (in red) and the second closest neighbors (pink) of *amazon* from 1990 to 2016 according to the NYT corpus. The top four words relate to the Amazon region of South America; this meaning remains stable throughout the years. Items related to the company (founded in 1994) begin to appear significantly in the corpus in 1998 and the two meanings (geography and business) are both present after 1998.

Apple's behavior is different, which is not surprising since the company was already in existence in 1990 (Table 2.) The two meanings coexist from the start, separated by *blackberry*, which can itself be a fruit or a device/company name (all appearances of *apple* in the New York Times refer to a fruit before 2000.)

| | '90 | '91 | '92 | '93 | '94 | '95 | '96 | '97 | '98 | '99 | '00 | '01 | '02 | '03 | '04 | '05 | '06 | '07 | '08 | '09 | '10 | '11 | '12 | '13 | '14 | '15 | '16 |
|---|---|---|---|---|---|---|---|---|---|---|---|---|---|---|---|---|---|---|---|---|---|---|---|---|---|---|---|
| cobbler | | | ■ | | | ■ | ■ | ■ | | ■ | ■ | ■ | | | | ■ | | | | | | | | | | | |
| charlotte | | | | | | | | | | | | | | | | | ▪ | | | | | | | | | | |
| tarte | | | | | ■ | | | | | | | | | ▪ | ▪ | ▪ | ▪ | ▪ | | | ▪ | | | ■ | | | |
| tartlet | | | | | | | | | | | | | | | | | | | | | | | | | ■ | | |
| blackberry | ■ | | | | ▪ | | | | | | | | | | | | | ■ | | | | ▪ | ■ | ▪ | ■ | ■ | ■ |
| compaq | ▪ | ■ | ■ | ■ | | | ■ | | | ▪ | | ▪ | ▪ | ▪ | | | | | | | | | | | | | |
| cupertino | | | ▪ | | | | | | | | | | | | | | | | | | ▪ | | | ▪ | | ▪ | ▪ |
| motorola | | | | | | | | | | | | | ▪ | | | | | ■ | | ▪ | | | ▪ | | ▪ | ▪ | |
| acer | | | | | | | ▪ | | ▪ | | | | | | | | | | | | | | | | | | |
| imac | | | | | | | | | | | | ■ | | ■ | | ■ | ■ | | | | | | ▪ | | ■ | | |
| in-app | | | | | | | | | | | | | | | | | | | | | | | | | | | ■ |
| Nr. of occurrences | 118 | 105 | 74 | 124 | 95 | 97 | 100 | 125 | 143 | 165 | 196 | 152 | 129 | 120 | 263 | 184 | 24 | 218 | 215 | 141 | 256 | 373 | 508 | 298 | 379 | 304 | 217 |

**Table 2 -- Time trajectory of *apple***
Apple was founded in 1976 and the two meanings (fruit and business) coexist from the start. *Blackberry* may refer either to the fruit or to the product which first appeared in 1999. There were very few mentions of the company in the New York Times between 2001 and 2005.

## 4. Semantic drift between epochs

To evaluate semantic drifts, we first built two semantic spaces, $\mathcal{S}_0 = \sum_k \mathcal{S}_{k \in [1990,\ldots 1995]}$ and $\mathcal{S}_1 = \sum_k \mathcal{S}_{k \in [2010,\ldots 2015]}$ covering the first and last six full years of the corpus by a linear combination of each year's semantic space (cf. Equation 2). The algorithm to evaluate the drift between (0) and (1) is then very simple:

**Algorithm 2** - Algorithm to evaluate the drift in meaning of terms between time periods $0$ and $1$

1 : Input: Semantic spaces $\mathcal{S}_0 = \sum_k \mathcal{S}_{k \in [1990,\ldots 1995]}$ and $\mathcal{S}_1 = \sum_k \mathcal{S}_{k \in [2010,\ldots 2015]}$ for the two time periods
2 : **for each** term $t$ in corpus
3 : Extract and normalize vectors $|\mathcal{T}_t^0\rangle$ and $|\mathcal{T}_t^1\rangle$ of $t$ from $\mathcal{S}_0$ and $\mathcal{S}_1$
4 : Compute $\sigma_{01} = \langle \mathcal{T}_t^0 | \mathcal{T}_t^1 \rangle$
5 : $\sigma_{01}$ is an indicator of the drift
6 : **return** The lists of terms closest to $v_0$ and $v_1$

This algorithm is very fast and can easily handle all the terms, but in practice a term must be frequent enough for the result to be significant; only the 5210 non-stop words occurring more than 1500 times have been kept.

Words can be grouped into four categories, as shown in Tables 3, 4, 5 and 6. In these four tables, the reference words are followed by the similarity $\sigma_{01}$ between their vectors $v_0$ and $v_1$ for the two time periods. The lists are of the words closest to $v_0$ and $v_1$ in each year and the numbers between parentheses are their total number of occurrences. The words underlined in blue are common between the two periods.

### a. Very stable words

Stable words exhibit a strong similarity ($\sigma_{01} \geqslant 0.70$) between periods (Table 3). This is the case for words such as *inches*, *feet*, *office*, *murder*, which show little change with time.

| | | |
|---|---|---|
| **inches** $\sigma_{01} = \mathbf{0.88}$ | 1990—95 | feet (5), pounds (4), diameter (3), crackle (1), bench-press (1), concours (1), cylinder (1), width (1), timbers (1), spacing (1), biceps (1), inline (1), metallurgist (1), coxes (1), hurdles (1), record-holder (1), emaciation (1), thickness (1) |
| | 2010—15 | feet (5), pounds (5), diameter (2), high-tops (1), implements (1), right-thinking (1), rainbow (1), tailors (1), aleksey (1), e-class (1), facemask (1), millville (1), danbury (1), clintonville (1), c-class (1), big-picture (1), longhi (1) |
| **office** $\sigma_{01} = \mathbf{0.85}$ | 1990—95 | state (6), house (4), district (4), yesterday (3), officials (3), money (3), department (3), percent (2), agency (2), city (2), services (2), states (1), government (1), deputy (1) |
| | 2010—15 | city (4), home (3), district (3), case (2), state (2), government (2), weeks (1), mukasey (1), part (1), lackawanna (1), deal (1), caroms (1), company (1), department (1), right (1), investigation (1) |
| **feet** $\sigma_{01} = \mathbf{0.77}$ | 1990—95 | inches (5), square (5), pounds (3), space (2), garden (1), madison (1), damage (1), flood (1), manezh (1), cul-de-sac (1), throwers (1), milliliter (1), red-bearded (1), rieger (1), swooned (1), calves (1) |
| | 2010—15 | inches (4), foot (4), pounds (4), square (2), body (2), washes (1), cuts (1), capacity (1), legs (1), two-putt (1), size (1), baths (1), tahrir (1), extra (1), force (1), squeeze (1), madison (1) |
| **murder** $\sigma_{01} = \mathbf{0.70}$ | 1990—95 | charges (4), trial (3), manslaughter (3), counts (3), criminal (2), case (2), kidnapping (2), defendants (2), prosecutors (2), lawyers (2), possession (1), rape (1), simpson (1), prosecution (1), riot (1) |
| | 2010—15 | counts (2), burglary (2), manslaughter (2), strangulation (2), robbery (2), prison (2), assadourian (1), benefactors (1), mother (1), rampart (1), frayer (1), shred (1), razzle-dazzle (1), sollecito (1), children (1), rape (1) |

**Table 3 -- Examples of stable words between 1990—1995 and 2010—2015**
For higher similarity values, despite some minor time-related shifts, word meanings show little significant change with time.

### b. Moderately stable words and food words

A second category is still reasonably stable and exhibits similarities $\sigma_{01} \approx 0.5$. This is the case for many food words shown in Table 4: eating habits have much changed over these 20 years, at least for readers of the *New York Times*: for example, the word *sauce* disappears in the second period[16]. In the first period, most of the words correspond to standard international fare but they become much more ethnic and more dispersed in the second period (note that apparently unrelated words may in fact be subtly related. For example, *sosinski* doesn't refer to the basketball player but to a Montauk lobster fisherman who almost drowned…)

| | | |
|---|---|---|
| **tomatoes** $\sigma_{01} =$ **0.62** | 1990—95 | peppers (6), tomato (5), mozzarella (4), garlic (4), sun-dried (4), ricotta (2), eggplant (2), olives (2), sauce (2), basil (2), onions (2), pesto (1), sauteed (1), salad (1) |
| | 2010—15 | cucumbers (3), tomato (2), romano (1), basil (1), bellota (1), pomegranate (1), marsala (1), tatin (1), escarole (1), garlic (1), pecorino (1), onion (1), tractor-trailer (1), endoscopic (1), sofrito (1), tarte (1), ricotta (1) |
| **salmon** $\sigma_{01} =$ **0.52** | 1990—95 | sauce (6), chicken (4), tuna (3), caviar (3), seared (2), tartars (2), salad (2), shrimp (1), sweet (1), spinach (1), lobster (1), sturgeon (1), creme (1), beans (1), fish (1), fraiche (1), goat-cheese (1), seafood (1), potatoes (1) |
| | 2010—15 | anchovies (2), herring (2), fish (1), naan (1), tilapia (1), nova (1), sardines (1), maguro (1), horseradish-crusted (1), recipe (1), tentacles (1), knowledgeably (1), sablefish (1), seafood (1), orzo (1), potvin (1), morsels (1), arrestees (1) |
| **potato** $\sigma_{01} =$ **0.47** | 1990—95 | salad (3), potatoes (2), chips (2), sauce (2), shrimp (2), spinach (2), tortilla (1), parsley (1), samosas (1), glenna (1), carrots (1), zucchini (1), cheez (1), thai-style (1), capita (1), pommes (1), once-a-week (1), pancakes (1), latticed (1), frites (1), fat-laden (1), elmer (1), renner (1) |
| | 2010—15 | ahoy (1), larking (1), starch (1), leeks (1), toffee (1), gluten-free (1), kmart (1), tyler (1), coconuts (1), chicken (1), cupboards (1), chaat (1), tikki (1), pulled-pork (1), aloo (1), tapioca (1), crisped (1), spuds (1), lovage (1), chickpeas (1) |
| **mushrooms** $\sigma_{01} =$ **0.40** | 1990—95 | garlic (4), sauce (4), peppers (4), mushroom (2), eggplant (2), sauteed (2), herbs (1), gusto (1), risotto (1), kachaloo (1), cheese (1), mariscada (1), duck (1), feather-light (1), pork (1), vinegar (1), shrimp (1), artichokes (1), gnocchi (1), tomato (1), portobello (1) |
| | 2010—15 | cremini (2), racchetta (1), demi-glace (1), portobello (1), garnish (1), strangling (1), scaloppine (1), racquetball (1), pizza (1), rainy-day (1), ricotta (1), escarole (1), shiitake (1), fruits (1), deliveryman (1), prosciutto (1), flavor-packed (1), chanterelle (1), marsala (1) |
| **seafood** $\sigma_{01} =$ **0.36** | 1990—95 | shrimp (4), salmon (3), sauteed (2), scallops (2), squid (2), tomato (2), mussels (2), lobster (2), sauce (2), clams (1), meat (1), misto (1), ravioli (1), vegetable (1), salad (1), pepper (1), garni (1), dishes (1), dover (1), communists (1), calamari (1), aioli (1) |
| | 2010—15 | mercury (1), hermanos (1), salmon (1), mariscos (1), bargain (1), wide-mouthed (1), oyster (1), greenpeace (1), tilapia (1), calle (1), shellfish (1), fortaleza (1), greenberg (1), dojo (1), arroz (1), publix (1), imported (1), traceability (1), fettuccine (1) |
| **lobster** $\sigma_{01} =$ **0.35** | 1990—95 | sauce (5), chicken (3), scallops (3), shrimp (3), bisque (2), consomme (1), meat (1), onions (1), clams (1), sweetbreads (1), tomatoes (1), rosemary (1), broth (1), squab (1), salmon (1), ravioli (1), tomato (1), red-wine (1), squid (1), salad (1) |
| | 2010—15 | sosinski (1), uneaten (1), midler (1), biriyani (1), dining-room (1), buoy (1), nuevo (1), vadouvan (1), duxelles (1), scratch (1), cloudlike (1), chowder (1), mushroom (1), stew (1), sogginess (1), bisque (1), ravioli (1), boat (1) |

**Table 4 -- Some food words between 1990—1995 and 2010—2015**
Despite similarities $\sigma_{01}$ being still midrange, the comparison shows that eating habits have much changed: note for example the complete disappearance of the word *sauce* between the two periods. Also, in the first period, most of the words correspond to standard international fare, while they become much more ethnic in the second period.

**c. Fast changing words**

The first four words in Table 5 illustrate that singular and plural forms of words may have very different evolutions: *prices* has about the same connotations in the two periods, while *price*, mostly related to restaurants and food items in the first period, becomes little different from *prices* in the second period.

Similarly, *devices* in the first period refers mostly to implants but converges with *device* in the second period, where it refers to technical devices, as did the singular form in the first period. $\sigma_{01}$ is nevertheless very low for *device* because, while still technical, the nature of the devices and their names have changed.

The last two words, *couples* and *marriage*, also strongly reflect the societal evolution of the last twenty years, with very low interperiod similarities.

| | | |
|---|---|---|
| **prices** $\sigma_{01} =$ **0.71** | 1990—95 | taxes (4), subject (4), market (4), percent (4), menu (3), listings (3), change (3), inflation (2), trading (1), reviews (1), industry (1), rise (1), ices (1), company (1), sales (1) |
| | 2010—15 | price (4), market (4), costs (3), percent (2), money (2), inflation (2), companies (2), economy (2), growth (1), emissions (1), means (1), payroll (1), rates (1), rate (1), government (1) |
| **price** $\sigma_{01} =$ **0.21** | 1990—95 | appetizers (5), range (4), entrees (3), courses (3), lunch (3), capsule (2), suburbs (2), symbols (2), restaurant (2), establishments (2), cuisine (2), taxes (2), reviews (2), dinner (2), location (2), inflation (1), prices (1), relation (1), desserts (1), market (1) |
| | 2010—15 | prices (5), company (4), market (4), shares (3), percent (3), value (2), investors (2), product (1), sales (1), investment (1), demand (1), size (1), game (1), klar (1), example (1), number (1), revenue (1), cost (1), money (1) |
| **devices** $\sigma_{01} =$ **0.20** | 1990—95 | silicone (2), implants (1), thumps (1), phase-change (1), scramblers (1), silicone-gel (1), saline (1), collaboration (1), attention (1), breast (1), testicles (1), bleeps (1), penis (1), catheters (1) |
| | 2010—15 | phones (3), apps (3), device (3), apple (2), users (2), android (2), analogue (1), microphone (1), software (1), amplification (1), suggests (1), persimmon (1), sleep (1), payments (1), augmentation (1), ikea (1), imovie (1), course (1), imac (1), smartphone (1), crier (1), tablets (1), form (1) |
| **device** $\sigma_{01} =$ **0.06** | 1990—95 | scientific-atlanta (1), portals (1), decompression (1), converter (1), chords (1), graphics (1), mcmorris (1), birth-control (1), tuner (1) |
| | 2010—15 | devices (3), tablet (1), social (1), users (1), supplies (1), cars (1), nominees (1), ubislate (1), system (1), brave (1), microprocessor (1), facebook (1), aakash (1), clearance (1), cedars (1), table (1) |
| **couples** $\sigma_{01} =$ **0.15** | 1990—95 | macbeth (1), incomes (1), retirees (1), marilyn (1), koreans (1), individuals (1) |
| | 2010—15 | marriage (5), heterosexual (2), licenses (2), lesbian (1), probate (1), institutionalization (1), childrearing (1), unwelcoming (1), cohabitating (1), weddings (1), carats (1), legalizes (1), none-too-relaxing (1), reunified (1), outserve (1), hertz (1), marriages (1), detriments (1), gregoire (1), households (1) |
| **marriage** $\sigma_{01} =$ **0.14** | 1990—95 | mendelsohn (1), admirer (1), malarkey (1), layabout (1), white-water (1), catullus (1), amiability (1), piers (1), storybook (1), pamela (1), middleton (1), chalone (1), fairy-tale (1), fiona (1) |
| | 2010—15 | couples (5), marriages (3), ellner (2), lesbians (2), licenses (2), puzzler (1), tailwind (1), same-sex (1), unwelcoming (1), poehler (1), malott (1), institutionalization (1), anathema (1), gregoire (1), grenell (1), non-orthodox (1), reunified (1), detriments (1), jurisprudential (1), heterosexual (1) |

**Table 5 -- Interesting evolutions between 1990—1995 and 2010—2015**
Note how evolution can be quite different for singular and plural forms of words, such as *price* and *prices* or *device* and *devices*, and how *couples* and *marriage* reflect the ongoing societal evolution.

### d. Unstable words

Some words (see Table 6) are inherently unstable because they refer to fast-changing items, such as *crisis*, *borders*, *moscow* or *muslims*. More surprising is the unstability of *lunch*, from eating food to a much less specific connotation.

| | | |
|---|---|---|
| **crisis** $\sigma_{01} = 0.29$ | 1990—95 | minesweepers (1), persian (1), boredom (1), world (1), action (1), villanueva (1), marshaled (1), midrock (1), deployment (1), discussions (1), support (1), conflict (1), buildup (1), oman (1), midlife (1), oil-rich (1), big-think (1), gulf (1), perennials (1), plenty (1), incursion (1), grammy (1) |
| | 2010—15 | financial (6), economy (5), government (4), banks (4), market (4), debt (2), institutions (2), rescue (1), recession (1), growth (1), information (1), default (1), industry (1), executives (1), contraction (1), deal (1), bailout (1), services (1), sovereign (1), company (1), greece (1), eurozone (1) |
| **borders** $\sigma_{01} = 0.28$ | 1990—95 | scrolled (1), yogi (1), riley (1), conacher (1), faceoff (1), economy (1), mccarter (1), chorske (1), phone (1), agreement (1) |
| | 2010—15 | doctors (2), group (1), guinea (1), core (1), fronti (1), kraig (1), waldron (1), syria (1), sprecher (1), curbside (1), cargo (1), country (1), europe (1), company (1), africa (1), ebola (1), migrants (1), countries (1), nurses (1), decins (1), potboiler (1), china (1), vidor (1), annick (1) |
| **moscow** $\sigma_{01} = 0.27$ | 1990—95 | russia (4), russian (4), yeltsin (2), boris (2), union (2), russians (1), estonia (1), latvia (1), criminality (1), blockade (1), megaproject (1), lithuania (1), leaders (1), gorbachev (1), entertainment (1), murmansk (1), diplomat (1), ukraine (1) |
| | 2010—15 | snowden (1), leaker (1), debaltseve (1), kvasha (1), ecuador (1), comstock (1), latif (1), peat (1), novosibirsk (1), putin (1), benfica (1), contractor (1), rozehnal (1), bradsher (1), datsyuk (1), russian (1), oleg (1), vladimir (1), crimea (1), spartak (1), pavel (1) |
| **muslims** $\sigma_{01} = 0.26$ | 1990—95 | serbs (4), croats (4), bosnian (3), muslim (2), realpolitik (1), territory (1), cleansing (1), najaf (1), non-arab (1), slavs (1), forefathers (1), sunni (1), saudi (1), kiseljak (1), croatia (1), croat (1), vares (1), slovenes (1), croatian (1), krajina (1), allots (1), christians (1), bosnia (1) |
| | 2010—15 | hindus (2), clerics (1), hindu (1), jats (1), wiccans (1), jihadist (1), amendola (1), mohammed (1), andhra (1), quran (1), torching (1), swamp (1), ahmadis (1), kumar (1), buddhists (1), fugana (1), sikhs (1) |
| **lunch** $\sigma_{01} = 0.18$ | 1990—95 | entrees (6), dinner (6), range (6), appetizers (6), courses (4), fixe (3), carte (2), desserts (2), prix-fixe (2), pre-theater (2), pastas (1), mondays (1), price (1), prix (1), noon (1), three-course (1), brunch (1) |
| | 2010—15 | fact (1), apartment (1), outing (1), wright-phillips (1), home (1), cronenberg (1), breakfast (1), meritocratic (1), dinner (1), bastille (1), ritz-carlton (1) |

**Table 6 -- Some unstable words**
*crisis*, *borders*, *moscow* and *muslims* have obviously evolved according to the international situation. The case of *lunch* is more surprising: it refers to a meal in the first period, but this connotation has almost disappeared in the second period.

## 5. Gender bias in the New York Times over the last twenty years

Garg *et al.* [10] have used a number of sources in their evaluation of gender bias and we have downloaded from their GitHub site [17] lists of adjectives of *appearance* and of *intelligence*, a list of *occupations*, and two lists of words related respectively to men and to women (*man terms* and *woman terms*, to use Garg *et al.* terminology).

Tables 7 and 8 have been constructed using the following algorithm:

---

**Algorithm 3** - Find most frequent men and women qualifiers in time periods $K_0$ and $K_1$

1 : Input: The lists of qualifiers and of man/woman terms
2 : **for each** year $k \in [K_0, K_1]$
3 :     **for each** qualifier $q$
4 :         Extract qualifier vectors $|\mathcal{Q}_q^k\rangle$ from *partial* semantic space $\mathcal{S}_k$
5 :         **for each** man term $m$
6 :             Extract male vectors $\mathcal{M}_m^k$ from $\mathcal{S}_k$
7 :             Compute $\sigma_{mq}^k = \langle \mathcal{M}_m^k | \mathcal{Q}_q^k \rangle$
8 :         **return** largest $\sigma_{mq}^k$
9 :         **for each** woman term $w$
10 :            Extract female vectors $|\mathcal{W}_w^k\rangle$ from $\mathcal{S}_k$
11 :            Compute $\sigma_{wq}^k = \langle \mathcal{W}_w^k | \mathcal{Q}_q^k \rangle$
12 :        **return** largest $\sigma_{wq}^k$
13 :        If $\sigma_{wq}^k > \sigma_{mq}^k$ qualifier $q$ is taken as preferentially female in year $k$ and conversely
14 :     **return** The gender attached to qualifier $q$ in year $k$
15 : **return** The lists of male and female qualifiers in time periods $K_0$ and $K_1$
16 : Create table of male and female qualifiers for both time periods

---

Given its readership, one would expect the New York Times to avoid gender stereotyping; this is indeed the case:

| | | |
|---|---|---|
| **1990 — 95** | male: | healthy, slim, attractive, slender, genius, ugly, brilliant, muscular, smart, homely, fashionable, judicious, resourceful |
| | female: | strong, beautiful, wise, thoughtful, intelligent, handsome, weak, fat, thin, inventive, blushing, clever, reflective, gorgeous |
| **2010 — 15** | male: | strong, healthy, ugly, brilliant, handsome, athletic, thoughtful, alluring, slim, clever, ingenious, imaginative, plump |
| | female: | intelligent, beautiful, smart, muscular, wise, attractive, logical, thin, fashionable, gorgeous, weak, genius, inventive, fat |

**Table 7 -- Top qualities by gender in the periods 1990 — 1995 and 2010 — 2015**
Apart perhaps from a few terms such as *ugly*, *beautiful* or *gorgeous*, gender differentiation is weak in both period and changes little over two decades.

| | | |
|---|---|---|
| **1990 — 95** | male: | police, official, doctor, secretary, lawyer, artist, driver, professor, manager, guard, athlete, physician, surgeon, dancer |
| | female: | author, teacher, baker, sales, pilot, farmer, soldier, cook, judge, designer, nurse, attendant, engineer, psychologist |
| **2010 — 15** | male: | police, official, guard, manager, soldier, architect, pilot, physician, driver, nurse, therapist, smith, mason, cook |
| | female: | doctor, lawyer, teacher, author, sales, designer, baker, professor, photographer, farmer, artist, operator, scientist, secretary |

**Table 8 -- Top occupations by gender in the periods 1990 — 1995 and 2010 — 2015**
Gender differentiation by occupation is also weak and conforms largely to well-established stereotypes; note also the interperiod shifts of *doctor*, *professor* or *lawyer* from male to female, and conversely of *nurse* from female to male. *police* and *official* remain stubbornly male in both periods.

# 6. Equivalences

While the concepts of *hurricane* or *president* are stable, their instantiations change with time. However, with the distributional hypothesis[11], we would expect most of their their neighboring vectors in different years to be close as they correspond to generic hurricanes or generic presidents. Table 9 was built with the following algorithm:

**Algorithm 4** - Algorithm to find in year $k$ the terms closest to term $t$ in year $k'$

1 : Input: All partial semantic spaces $\mathcal{S}_k$
2 : Extract and normalize vector $|\mathcal{T}_t^{k'}\rangle$ of term $t$ from the *partial* semantic space $\mathcal{S}_{k'}$
3 : **for each** year $k \in K$
4 : **return** the two closest neighbors of $|\mathcal{T}_t^{k'}\rangle$ in the *partial* semantic space $\mathcal{S}_k$
5 : Tabulate top two neighbors

The result is shown in Table 9: the vectors of an instantiation in the relevant year $k'$ from semantic space $\mathcal{S}_{k'}$ (say *katrina$_{2005}$* or *obama$_{2016}$*) are indeed most often closest to each year's instantiation: for example, *andrew$_{1992}$* or *floyd$_{1999}$* are closest to *katrina$_{2005}$* etc.

Note that *andrew$_{1993}$* is only second closest to *katrina$_{2005}$* and comes after *fanfare*. This is due to the fact that in this partition actual fanfares are associated with proper names such as *andrew*: this is an artefact of the distributional hypothesis, which is also one of the reasons why the catastrophic hurricane *katrina* casts a long shadow after 2005: the term does not necessarily refer to the hurricane, both before and after 2005, and this introduces some uncertainty. Similarly, *bush* or *clinton* do not always refer to a president, or *hillary* to Ms. Clinton (her first appearance in this New York Times corpus dates from 1992.) Even *clinton* in 1990 refers less than 50% of the time to Arkansas Governor and future U.S. President Bill Clinton.

| Terms equivalent to *katrina$_{2005}$* | | | Terms equivalent to *obama$_{2016}$* | | Terms equivalent to *clinton$_{1996}$* | | Terms equivalent to *hillary$_{2015}$* | | |
|---|---|---|---|---|---|---|---|---|---|
| Year | Top term | Next term | Top term | Next term | Top term | Next term | Top term | Next term | Year |
| **1990** | — | — | **bush** | sort | marcos | edelman | darren | ucla | **1990** |
| **1991** | segment | acres | **bush** | took | **bush** | number | elephant | wesley | **1991** |
| **1992** | **andrew** | tasks | **bush** | campaign | **bush** | **clinton** | hillary | **bill** | **1992** |
| **1993** | fanfare | **andrew** | country | television | **clinton** | support | hillary | **administration** | **1993** |
| **1994** | — | — | **clinton** | support | **clinton** | support | hillary | mack | **1994** |
| **1995** | **felix** | **luis** | **clinton** | resident | **clinton** | campaign | hillary | **president** | **1995** |
| **1996** | hurricane | storms | **clinton** | president | **clinton** | campaign | hillary | **president** | **1996** |
| **1997** | — | — | **clinton** | violence | **clinton** | campaign | hillary | **president** | **1997** |
| **1998** | **bonnie** | wilmington | candidates | visitors | **clinton** | number | hillary | **president** | **1998** |
| **1999** | **floyd** | floods | **clinton** | congress | **clinton** | campaign | hillary | **clinton** | **1999** |
| **2000** | — | — | business | message | **clinton** | need | hillary | **clinton** | **2000** |
| **2001** | — | — | campaign | **bush** | **bush** | **clinton** | hillary | **rodham** | **2001** |
| **2002** | — | — | **bush** | supporters | **bush** | others | hillary | mama | **2002** |
| **2003** | **isabel** | wings | **bush** | governor | **bush** | support | hillary | mama | **2003** |
| **2004** | **ivan** | **frances** | **obama** | **clinton** | congress | **clinton** | hillary | crowd | **2004** |
| **2005** | **katrina** | **rita** | candidates | **bush** | **bush** | government | hillary | wounds | **2005** |
| **2006** | **katrina** | category | message | masters | column | study | hillary | ring | **2006** |
| **2007** | **katrina** | tornado | **obama** | illinois | **bush** | democrats | hillary | **obama** | **2007** |
| **2008** | **katrina** | orleans | **obama** | president-elect | **clinton** | **bush** | hillary | **senators** | **2008** |
| **2009** | — | — | **obama** | michelle | **obama** | **clinton** | hillary | **clinton** | **2009** |
| **2010** | — | — | **obama** | campaign | **clinton** | **obama** | hillary | **clinton** | **2010** |
| **2011** | category | larkin | **obama** | **bush** | **clinton** | survival | hillary | **clinton** | **2011** |
| **2012** | **katrina** | **irene** | **obama** | something | **clinton** | government | hillary | **secretary** | **2012** |
| **2013** | **katrina** | aftermath | **obama** | something | **obama** | **clinton** | hillary | blazer | **2013** |
| **2014** | **katrina** | aftermath | **obama** | home | **obama** | **clinton** | hillary | predecessor | **2014** |
| **2015** | **katrina** | **joaquin** | **obama** | candidate | **clinton** | **obama** | hillary | **secretary** | **2015** |
| **2016** | blessing | category | **obama** | hillary | **clinton** | **obama** | hillary | **secretary** | **2016** |

Table 9 -- Top two closest terms to *katrina$_{2005}$*, *obama$_{2016}$*, *clinton$_{1996}$* and *hillary$_{2015}$* from 1990 to 2016

The empty lines in the *katrina* column correspond to years without significant hurricanes (i.e. class 2 and higher), according to the National Hurricane Center reports[18]. Terms in bold are the correct answers but other terms are often quite relevant (*illinois* for *obama*, or *campaign* for both presidents.) The year 2006 should be ignored as this partition is an outlier with a much shorter vocabulary.

# 7. Order encoding and positional neighbors

As shown by Sahlgren et al.[19], permutations are a means to encode order in word space (see also Widdows et al.[13].)

Let's define permutation $\Pi^0$ as the natural sequence of the first $d$ integers,

$$\Pi^0 = (0, 1, 2, \ldots d-1) \qquad (3)$$

and $\Pi^1$ as a random permutation of the first $d$ integers, then $\Pi^{-1}$, $\Pi^2$ and $\Pi^{-2}$ are defined by

$$\Pi^{-1} \times \Pi^1 \equiv \Pi^1 \times \Pi^{-1} = \Pi^0, \qquad \Pi^2 = \Pi^1 \times \Pi^1 \qquad \text{and} \qquad \Pi^{-2} = \Pi^{-1} \times \Pi^{-1} \qquad (4)$$

If we randomly select a vector $|\mathcal{T}\rangle$ in a space of large enough dimension $d$, the vector $|\mathcal{V}\rangle = \Pi^1|\mathcal{T}\rangle$ obtained by permutation of the indices will clearly be quasi-orthogonal[19] to $|\mathcal{T}\rangle$, provided $\Pi^1$ has no fixed point.

If we replace Equation 1 with the following:

$$|\mathcal{V}_i\rangle = \Pi^{-2}(\rho_{i-2}|\mathcal{T}_{i-2}^s\rangle) + \Pi^{-1}(\rho_{i-1}|\mathcal{T}_{i-1}^s\rangle) + \Pi^0(\rho_i|\mathcal{T}_i^s\rangle) + \Pi^1(\rho_{i+1}|\mathcal{T}_{i+1}^s\rangle) + \Pi^2(\rho_{i+2}|\mathcal{T}_{i+2}^s\rangle) \qquad (5)$$

where $|\mathcal{T}_{i+\delta}^s\rangle$ is the normalized seed vector of term $t_{i+\delta}$ and $\rho_{i+\delta}$ is, as above, a multiplicative coefficient (for example a weight) the scalar product of (unnormalized) $|\mathcal{V}_i\rangle$ with the permuted, normalized seed vector $\Pi^\delta|\mathcal{T}_{i+\delta}^s\rangle$ will yield $\rho_{i+\delta} + \epsilon$ where $\epsilon$ is a zero-centered noise contribution. This fact can be used to retrieve the most likely immediate positional neighbors of any word, as long as the noise remains acceptable[20].

| | Predict word in next position | | | | | Predict words in previous and next positions | |
|---|---|---|---|---|---|---|---|
| president ? | york ? | columbia ? | supreme ? | world ? | year | ? trade ? | ? states ? |
| president **bush** | york **city** | columbia **university** | supreme **court** | world **war** | 1990 | **national** trade **center** | **united** states **government** |
| president **bush** | york **city** | columbia **university** | supreme **court** | world **war** | 1991 | **national** trade **center** | **united** states **american** |
| president **bush** | york **city** | columbia **university** | supreme **court** | world **war** | 1992 | **national** trade **center** | **united** states **government** |
| president **clinton** | york **city** | columbia **university** | supreme **court** | world **war** | 1993 | **world** trade **center** | **united** states **government** |
| president **clinton** | york **city** | columbia **university** | supreme **court** | world **war** | 1994 | **national** trade **center** | **united** states **government** |
| president **clinton** | york **city** | columbia **university** | supreme **court** | world **war** | 1995 | **world** trade **center** | **united** states **government** |
| president **clinton** | york **city** | columbia **university** | supreme **court** | world **war** | 1996 | **national** trade **center** | **united** states **party** |
| president **clinton** | york **times** | columbia **university** | supreme **court** | world **war** | 1997 | **world** trade **center** | **united** states **government** |
| president **clinton** | york **times** | columbia **university** | supreme **court** | world **series** | 1998 | **world** trade **center** | **united** states **government** |
| president **clinton** | york **times** | columbia **university** | supreme **court** | world **war** | 1999 | **world** trade **center** | **united** states **party** |
| president **clinton** | york **times** | columbia **university** | supreme **court** | world **series** | 2000 | **world** trade **center** | **united** states **party** |
| president **bush** | york **times** | columbia **university** | supreme **court** | world **trade** | 2001 | **world** trade **center** | **united** states **american** |
| president **bush** | york **times** | columbia **university** | supreme **court** | world **war** | 2002 | **world** trade **center** | **united** states **american** |
| president **bush** | york **times** | columbia **university** | supreme **court** | world **war** | 2003 | **world** trade **center** | **united** states **american** |
| president **bush** | york **city** | columbia **university** | supreme **court** | world **war** | 2004 | **world** trade **center** | **united** states **american** |
| president **bush** | york **city** | columbia **university** | supreme **court** | world **war** | 2005 | **world** trade **center** | **united** states **american** |
| president **clinton** | york **times** | columbia **sunday** | supreme **village** | world **theater** | 2006 | **east** trade **street** | **united** states **park** |
| president **bush** | york **city** | columbia **university** | supreme **court** | world **war** | 2007 | **world** trade **center** | **united** states **american** |
| president **obama** | york **city** | columbia **university** | supreme **court** | world **war** | 2008 | **world** trade **obama** | **united** states **american** |
| president **obama** | york **times** | columbia **university** | supreme **court** | world **war** | 2009 | **senator** trade **obama** | **united** states **government** |
| president **obama** | york **times** | columbia **university** | supreme **court** | world **series** | 2010 | **world** trade **center** | **united** states **party** |
| president **obama** | york **times** | columbia **university** | supreme **court** | world **war** | 2011 | **world** trade **center** | **united** states **government** |
| president **obama** | york **city** | columbia **university** | supreme **court** | world **series** | 2012 | **world** trade **obama** | **united** states **government** |
| president **obama** | york **city** | columbia **university** | supreme **court** | world **war** | 2013 | **world** trade **center** | **united** states **government** |
| president **obama** | york **city** | columbia **university** | supreme **court** | world **war** | 2014 | **world** trade **center** | **united** states **government** |
| president **obama** | york **city** | columbia **university** | supreme **court** | world **series** | 2015 | **world** trade **center** | **united** states **government** |
| president **obama** | york **city** | columbia **university** | supreme **court** | world **war** | 2016 | **world** trade **center** | **united** states **party** |

Table 10 -- Positional neighbors of selected words
Except for the year 2006, which should be ignored as it is an outlier with a much shorter vocabulary, the guessed words are coherent with what would be expected.

This is shown in Table 10 : ignoring outlying year 2006, the term following *president* is almost always the name of the actual president in the relevant year, *york* is followed by either *times* or *city*, *supreme* is always followed by *court*, *series* and *war* may both occur after *world*. As should be expected, predictions are less effective in the trigram situation since, for example, the digrams *world trade* and *trade center* do occur in other trigrams than *world trade center* which occurs 77% of the time; *united states government* occurs only half the time because *united states* occurs much more frequently (47,463 times vs. 673 times.)

## 8. Conclusions

The linearity of random vectors techniques enables the building of time series of semantic spaces which can then be compared and combined linearly without loss of precision over any time period to detect diachronic semantic shifts. In this article, we have demonstrated several uses of this approach: visualization of time trajectories of polysemous words such as *amazon* or *apple*, detection of semantic drifts and of gender bias across time, revealing the shifting instantiations of stable concepts such as *hurricane* or *president*. This very fast, linear approach can easily be distributed over many processors to follow real time language evolutions and the resulting, time-dependent semantic spaces can then be combined at will by simple additions or subtractions. This being a statistical process, its time resolution will be dependent on sample size and thus on the volume (number of words) of each time slice: one year is a lower limit for the sample of the New York Times [7] we used but new Wikipedia downloads can be followed as they are posted, new patents and innovations can be followed on a weekly basis and streams of social media such as those of Twitter or Facebook could probably be followed with a time resolution of a few seconds.

## Bibliography

[1] **Mikolov, T., Chen, K., Corrado, G. and Dean, J.**, Efficient Estimation of Word Representations in Vector Space, *arXiv:1301.3781v3*, 2013.

[2] **Mikolov, T., Sutskever, I., Chen, K., Corrado, G. and Dean, J.**, Distributed Representations of Words and Phrases and their Compositionality, *arXiv:1310.4546v1*, 2013.

[3] **Goldberg, Y. and Levy, O.**, Neural Word Embedding as Implicit Matrix Factorization, In *Advances in Neural Information Processing Systems 27 (NIPS)*, 2014.

[4] **Levy, O., Søgaard, A., Goldberg, Y.**, A Strong Baseline for Learning Cross-Lingual Word Embeddings from Sentence Alignments, *arXiv:1608.05426v2 [cs.CL]*, 2017.

[5] **Arora, S.** Word Embeddings: Explaining their properties, http://www.offconvex.org/2016/02/14/word-embeddings-2/ Downloaded on May 8, 2018.

[6] **Yao, Z., Sun, Y., Ding, W., Rao, N., and Xiong, H.**, Dynamic Word Embeddings for Evolving Semantic Discovery, In WSDM 2018: The Eleventh ACM International Conference on Web Search and Data Mining, February 5 — 9, 2018, *arXiv:1703.00607v2*, Feb. 13, 2018.

[7] **Yao, Z., Sun, Y., Ding, W., Rao, N., and Xiong, H.**, The data is available at https://sites.google.com/site/zijunyaorutgers/ and was downloaded on February 23, 2018.

[8] **Kulkarni, V., Al-Rfou, R., Perozzi, B., and Skiena, S.**, Statistically Significant Detection of Linguistic Change, *arXiv:1411.3315v1*, Nov. 12, 2014.

[9] **Hamilton, W., Leskovec, J., and Jurafsky, D.** Diachronic Word Embeddings Reveal Statistical Laws of Semantic Change, *arXiv:1605.09096v4 [cs.CL]*, September 2016.

[10] **Garg, N., Schiebinger, L., Jurafsky, D. and Zou, J.** Word embeddings quantify 100 years of gender and ethnic stereotypes, Proceedings of the National Academy of Sciences (PNAS), March 2018.
Data and code related to this paper are available on GitHub
https://github.com/nikhgarg/EmbeddingDynamicStereotypes (downloaded April 17, 2018).

[11] **Firth, J. R.** stated that "You shall know a word by the company it keeps", in A Synopsis of Linguistic Theory, 1930-1955,
*Studies in Linguistic Analysis, Special volume of the Philological Society, Oxford, UK*, 1962.

[12] **Sahlgren, M.**, The Word-Space Model: Using Distributional Analysis to Represent Syntagmatic and Paradigmatic Relations between Words in High-dimensional Vector Spaces, *PhD Dissertation*, Stockholm University, Sweden, 2006.

[13] **Widdows, D. and Cohen, T.**, The Semantic Vectors Package: New Algorithms and Public Tools for Distributional Semantics, *Fourth IEEE International Conference on Semantic Computing (IEEE ICSC2010)*, 2010.

[14] **Delpech, J.-F.**, Unsupervised word sense disambiguation in dynamic semantic spaces, *arXiv:1802.02605v2*, Feb. 8, 2018.

[15] **Delpech, J.-F.** Semantic Technology-Assisted Review (STAR): Document analysis and monitoring using random vectors, *arXiv:1711.10307*, 2017.

[16] The fact that *sauce* disappears from the neighborhood of food terms doesn't mean that the term has disappeared (it is in fact only slightly less frequent in the second epoch) but simply that it occurs less frequently within windows containing the reference term.

[17] After Garg *et al.* [10], the following terms have been used to check stereotypes:
- Appearance terms such as *alluring*, *voluptuous*, *homely*, *muscular*… from
  https://github.com/nikhgarg/EmbeddingDynamicStereotypes/blob/master/data/adjectives_appearance.txt
- Intelligence terms such as *precocious*, *inventive*, *adaptable*, *reflective*… from
  https://github.com/nikhgarg/EmbeddingDynamicStereotypes/blob/master/data/adjectives_intelligencegeneral.txt
- Occupation terms such as *judge*, *dancer*, *optometrist*… from
  https://github.com/nikhgarg/EmbeddingDynamicStereotypes/blob/master/data/occupation_percentages_gender_occ1950.csv
- Terms exclusively relating to men such as *son*, *his*, *him*, *father*, *man*… from
  https://github.com/nikhgarg/EmbeddingDynamicStereotypes/blob/master/data/male_pairs.txt
- Terms exclusively relating to women such as *daughter*, *mother*, *woman*, *girl*… from
  https://github.com/nikhgarg/EmbeddingDynamicStereotypes/blob/master/data/female_pairs.txt

[18] NOAA—HRS Complete list of continental U.S. landfalling hurricanes
http://www.aoml.noaa.gov/hrd/tcfaq/E23.html downloaded on April 16, 2018.

[19] **Sahlgren, M., Holst, A., Kanerva, P.**, Permutations as a Means to Encode Order in Word Space
*Proceedings of the 30th Annual Meeting of the Cognitive Science Society (CogSci'08), Washington D.C.*, 2008.

[20] In this experiment, we use vectors of dimension 300 while Sahlgren *et al.* [19] use vectors are of dimension 3000, which should reduce noise by a factor $\sqrt{10}$.